# SINGULAR CURVES IN THE JOINT SPACE AND CUSP POINTS OF *3-RPR* PARALLEL MANIPULATORS

Mazen ZEIN, Philippe Wenger and Damien Chablat

Institut de Recherche en Communications et Cybernétique de Nantes UMR CNRS 6597

1, rue de la Noë, BP 92101, 44312 Nantes Cedex 03 France

E-mail address: Mazen.Zein@irccyn.ec-nantes.fr

Phone: 0033240376955    Fax: 0033240376930

**Abstract**

*This paper investigates the singular curves in the joint space of a family of planar parallel manipulators. It focuses on special points, referred to as cusp points, which may appear on these curves. Cusp points play an important role in the kinematic behavior of parallel manipulators since they make possible a nonsingular change of assembly mode. The purpose of this study is twofold. First, it exposes a method to compute joint space singular curves of* 3-R*P*R *planar parallel manipulators. Second, it presents an algorithm for detecting and computing all cusp points in the joint space of these same manipulators.*

## 1. Introduction

Because at a singularity a parallel manipulator loses its stiffness, it is of primary importance to be able to characterize these special configurations. This is, however, a very challenging task for a general parallel manipulator[1]. Planar parallel manipulators have received a lot of attention[1-8] because of their relative simplicity with respect to their spatial counterparts. Moreover, studying the former may help understand the latter. Planar manipulators with three extensible leg rods, referred to as 3-*RPR* manipulators, have often been studied. Such manipulators may have up to six assembly modes[7]. The direct kinematics can be written in a polynomial of degree six[3]. Moreover, as is the case in most parallel manipulators, the singularities coincide with the set of configurations where two direct kinematic solutions coincide. It was first pointed out that to move from one assembly mode to another, the manipulator should cross a singularity[7]. However, Innocenti[8] showed, using numerical experiments, that this statement is not true in general. More precisely, this statement is only true under some special geometric conditions, such as similar base and mobile platforms[9,10]. In fact, an analogous phenomenon exists in serial manipulators that can change their posture (inverse kinematic solution) without meeting a singularity in general, but not under special geometric simplifications[8,11-13]. The nonsingular change of posture in serial manipulators was shown to be associated with the existence of points in the workspace where three inverse kinematic solutions meet, called cusp points[13]. Cusp points in serial manipulators were determined by looking for the triple roots of the inverse kinematics polynomial[13] or from the equation of the workspace boundary[14]. Likewise, McAree[9] pointed out that for a 3-R*P*R parallel manipulator (as well as for its spatial counterpart, the octahedral manipulator), if a point with triple direct kinematic solutions exists in the joint space, then the nonsingular change of assembly mode is possible. A condition for three direct kinematic solutions to coincide was established. However, no systematic exploitation of this condition was possible because the algebra involved was too complicated and to the authors' knowledge, the work of McAree[9] has never been pursued yet. Wenger[15] showed that to accomplish a non-singular assembly-mode changing motion, a *3-RPR* manipulator platform should encircle a cusp point in its joint space. Thus, determination of cusp points is of interest for planning trajectories. A procedure for computing joint space singularities of *3-RPR* parallel manipulators was established in a previous paper[16], where the cusp points were shown on the joint space singularities of these manipulators but no solution was proposed for computing these cusp points.

In this paper, the abovementioned condition for computing cusp points is reviewed and exploited. An algorithm for the systematic detection of cusp points is developed. The method[16] for computing and representing joint space singularities of 3-R*P*R planar parallel manipulators is first recalled. A descriptive analysis of the singular curves in slices of the joint space of 3-R*P*R parallel manipulators is conducted. It is shown that the number of cusp points depends on the slice of the joint space in which these cusp points are determined. Moreover, the maximum number of cusp points depends on the geometry of the manipulator. This work helps better understand the topology of the joint space of parallel manipulators and finds applications in both design and trajectory planning.

The following section introduces the 3-R$\underline{P}$R manipulator and its constraint equations. Section 3 is devoted to the determination of the singular curves in slices of the joint space. The existence condition of cusp points is derived in section 4 and an algorithm to determine these points automatically is provided. Section 6 is devoted to a descriptive analysis of singular curves in slices of the joint space.

## 2. Preliminaries

### 2.1 Manipulators under study

The manipulators under study are 3-DOF planar parallel manipulators with three extensible leg rods (Fig. 1). These manipulators have been frequently studied[4-8]. Each of the three extensible leg rods is actuated with a prismatic joint. The geometric parameters of the manipulators are the three sides of the moving platform $d_1$, $d_2$, $d_3$ and the position of the base revolute joint centers defined by $A_1$, $A_2$ and $A_3$. The reference frame is centered at $A_1$ and the x-axis passes through $A_2$. Thus, $A_1 = (0, 0)$, $A_2 = (A_{2x}, 0)$ and $A_3 = (A_{3x}, A_{3y})$.

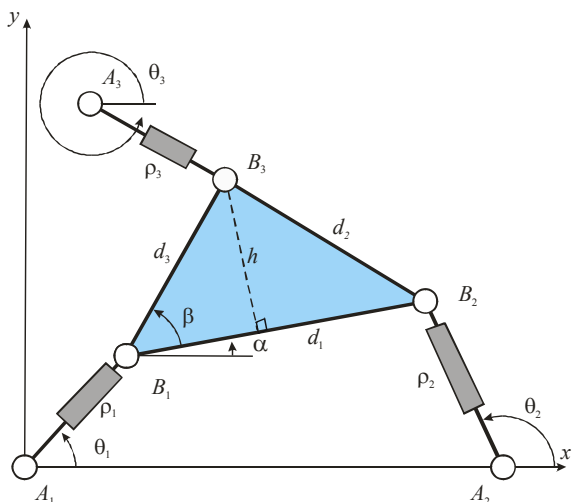

*Fig 1. The 3-R$\underline{P}$R parallel manipulator under study.*

## 2.2 Constraint equations

Let $\mathbf{L} \equiv (\rho_1, \rho_2, \rho_3)$ define the lengths of the three leg rods and let $\mathbf{\theta} \equiv (\theta_1, \theta_2, \theta_3)$ define the three angles between the leg rods and the *x*-axis. The six parameters $(\mathbf{L}, \mathbf{\theta})$ can be regarded as a configuration of the manipulator but only three of them are independent, so that the configuration space is a 3-dimensional manifold embedded in a 6-dimensional space. The dependency between $(\mathbf{L}, \mathbf{\theta})$ can be identified by writing the fixed distances between the three vertices of the mobile platform $B_1$, $B_2$, $B_3$:

$$\begin{cases} \Gamma_1(\mathbf{L},\mathbf{\theta}) = [\mathbf{b}_2(\mathbf{L},\mathbf{\theta}) - \mathbf{b}_1(\mathbf{L},\mathbf{\theta})]^T [\mathbf{b}_2(\mathbf{L},\mathbf{\theta}) - \mathbf{b}_1(\mathbf{L},\mathbf{\theta})] - d_1^2 = 0 \\ \Gamma_2(\mathbf{L},\mathbf{\theta}) = [\mathbf{b}_3(\mathbf{L},\mathbf{\theta}) - \mathbf{b}_2(\mathbf{L},\mathbf{\theta})]^T [\mathbf{b}_3(\mathbf{L},\mathbf{\theta}) - \mathbf{b}_2(\mathbf{L},\mathbf{\theta})] - d_2^2 = 0 \\ \Gamma_3(\mathbf{L},\mathbf{\theta}) = [\mathbf{b}_1(\mathbf{L},\mathbf{\theta}) - \mathbf{b}_3(\mathbf{L},\mathbf{\theta})]^T [\mathbf{b}_1(\mathbf{L},\mathbf{\theta}) - \mathbf{b}_3(\mathbf{L},\mathbf{\theta})] - d_3^2 = 0 \end{cases} \quad (1)$$

where $\mathbf{b}_i$ is the vector defining the coordinates of $B_i$ in the reference frame as function of $\mathbf{L}$ and $\mathbf{\theta}$. For more simplicity, $(\mathbf{L}, \mathbf{\theta})$ will be omitted in the following equations.

Expanding each $\Gamma_i$ as a series about the configuration $(\mathbf{L}, \mathbf{\theta})$ yields

$$\Delta\Gamma_i = \left( \sum_{j=1}^{3} \Delta\theta_i \frac{\partial}{\partial \theta_j} + \sum_{j=1}^{3} \Delta\rho_i \frac{\partial}{\partial \rho_j} \right) \Gamma_i + \frac{1}{2!} \left( \sum_{j=1}^{3} \Delta\theta_i \frac{\partial}{\partial \theta_j} + \sum_{j=1}^{3} \Delta\rho_i \frac{\partial}{\partial \rho_j} \right)^2 \Gamma_i + \ldots + \frac{1}{n!} \left( \sum_{j=1}^{3} \Delta\theta_i \frac{\partial}{\partial \theta_j} + \sum_{j=1}^{3} \Delta\rho_i \frac{\partial}{\partial \rho_j} \right)^n \Gamma_i + \ldots = 0 \quad (2)$$

If one keeps only the first-order and second-order terms, Eq.(2) can be written in matrix form as follows:

$$\Delta\mathbf{\Gamma} = \frac{\partial \mathbf{\Gamma}}{\partial \mathbf{\theta}} \Delta\mathbf{\theta} + \frac{\partial \mathbf{\Gamma}}{\partial \mathbf{L}} \Delta\mathbf{L} + \frac{1}{2} \begin{bmatrix} \Delta\mathbf{\theta}^T \frac{\partial^2 \Gamma_1}{\partial \mathbf{\theta}^2} \Delta\mathbf{\theta} \\ \Delta\mathbf{\theta}^T \frac{\partial^2 \Gamma_2}{\partial \mathbf{\theta}^2} \Delta\mathbf{\theta} \\ \Delta\mathbf{\theta}^T \frac{\partial^2 \Gamma_3}{\partial \mathbf{\theta}^2} \Delta\mathbf{\theta} \end{bmatrix} + \begin{bmatrix} \Delta\mathbf{\theta}^T \frac{\partial^2 \Gamma_1}{\partial \mathbf{\theta} \partial \mathbf{L}} \Delta\mathbf{L} \\ \Delta\mathbf{\theta}^T \frac{\partial^2 \Gamma_2}{\partial \mathbf{\theta} \partial \mathbf{L}} \Delta\mathbf{L} \\ \Delta\mathbf{\theta}^T \frac{\partial^2 \Gamma_3}{\partial \mathbf{\theta} \partial \mathbf{L}} \Delta\mathbf{L} \end{bmatrix} + \frac{1}{2} \begin{bmatrix} \Delta\mathbf{L}^T \frac{\partial^2 \Gamma_1}{\partial \mathbf{L}^2} \Delta\mathbf{L} \\ \Delta\mathbf{L}^T \frac{\partial^2 \Gamma_2}{\partial \mathbf{L}^2} \Delta\mathbf{L} \\ \Delta\mathbf{L}^T \frac{\partial^2 \Gamma_3}{\partial \mathbf{L}^2} \Delta\mathbf{L} \end{bmatrix} = 0 \quad (3)$$

Equation (3) can be used to describe an arbitrary local motion at a given configuration of the manipulator[9]. When first order terms of Eq.(3) are sufficient to describe the motion, the manipulator is in a regular configuration and the following equation (4) can be used instead of Eq.(3):

$$\frac{\partial \Gamma(\mathbf{L},\boldsymbol{\theta})}{\partial \boldsymbol{\theta}} \Delta \boldsymbol{\theta} + \frac{\partial \Gamma(\mathbf{L},\boldsymbol{\theta})}{\partial \mathbf{L}} \Delta \mathbf{L} = 0 \qquad (4)$$

Otherwise the configuration ($\mathbf{L},\boldsymbol{\theta}$) is special and the manipulator meets a singularity. This happens when the constraint Jacobian $\partial \Gamma / \partial \boldsymbol{\theta}$ drops rank so that the second order terms of equation (3) are required to describe the constraints. The three vertices of the moving platform have the following coordinates in the fixed reference frame,

$$\mathbf{b}_1 = \begin{bmatrix} \rho_1 \cos(\theta_1) & \rho_1 \sin(\theta_1) \end{bmatrix}^T ; \mathbf{b}_2 = \begin{bmatrix} A_{2x} + \rho_2 \cos(\theta_2) & \rho_2 \sin(\theta_2) \end{bmatrix}^T ;$$
$$\mathbf{b}_3 = \begin{bmatrix} A_{3x} + \rho_3 \cos(\theta_3) & A_{3y} + \rho_3 \sin(\theta_3) \end{bmatrix}^T .$$

Thus, the constraint Jacobian can be put in the following form:

$$\frac{\partial \Gamma}{\partial \boldsymbol{\theta}} = 2 \begin{bmatrix} \rho_1(A_{2x}s_1 + \rho_2 s_{12}) & \rho_2(\rho_1 s_{21} - A_{2x}s_2) & 0 \\ 0 & \begin{matrix} -\rho_2((A_{2x}-A_{3x})s_2 \\ -\rho_3 s_{23} + A_{3y}c_2) \end{matrix} & \begin{matrix} \rho_3((A_{2x}-A_{3x})s_3 \\ -\rho_2 s_{23} + A_{3y}c_3) \end{matrix} \\ \begin{matrix} \rho_1(A_{3x}s_1 - \rho_3 s_{31} \\ -A_{3y}c_1) \end{matrix} & 0 & \begin{matrix} -\rho_3(A_{3x}s_3 - \rho_1 s_{31} \\ -A_{3y}c_3) \end{matrix} \end{bmatrix} \qquad (5)$$

where $s_i = \sin(\theta_i)$, $c_i = \cos(\theta_i)$ and $s_{ij} = \sin(\theta_i - \theta_j)$.

3. **Jointspace Singular Curves**

The manipulator is in a singular configuration whenever the axes of its three leg rods are concurrent or parallel[17].

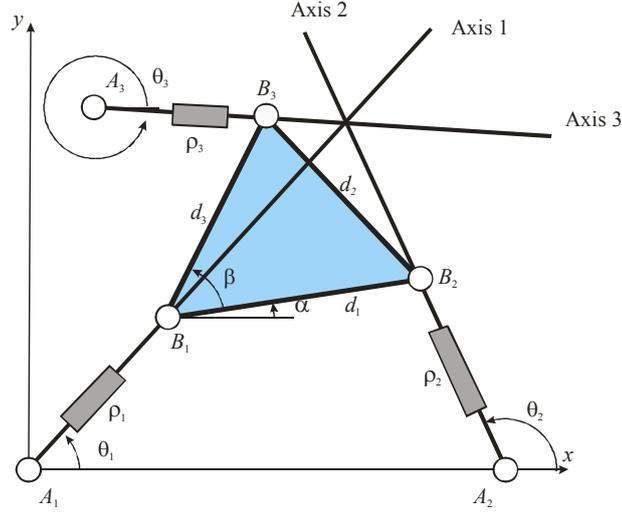

Fig 2. *A 3-RPR parallel manipulator on a singular configuration*

We derive the condition for the three leg axes to intersect at a common point (possibly at infinity). We first write the equations of the three leg axes:

$$\begin{cases} (\text{Axis 1}) : y\cos(\theta_1) = x\sin(\theta_1) \\ (\text{Axis 2}) : y\cos(\theta_2) = (x - A_{2x})\sin(\theta_2) \\ (\text{Axis 3}) : y\cos(\theta_3) = (x - A_{3x})\sin(\theta_3) + A_{3y}\cos(\theta_3) \end{cases} \quad (6)$$

Eliminating *x* and *y* in (6) yields the following singularity equation in the task parameters ($\theta_1$, $\theta_2$, $\theta_3$):

$$A_{2x}s_2 s_{31} + \left(A_{3x}s_3 - A_{3y}c_3\right)s_{12} = 0 \quad (7)$$

where $s_i = \sin(\theta_i)$, $c_i = \cos(\theta_i)$ and $s_{ij} = \sin(\theta_i - \theta_j)$.

It is possible to express Eq (7) as function of the joint space parameters $\rho_1$, $\rho_2$ and $\rho_3$ by using the constraint equations of the *3-RPR* manipulator. However, the resulting equation would be too complicated to yield real insights and difficult to handle. Our approach to compute the singular configurations in the joint space consists in reducing the dimension of the problem. We consider two-dimensional slices of the configuration space by fixing the first leg rod length $\rho_1$.

***Step 1:*** We rewrite Eq. (7) as a function of $\rho_1$, $\alpha$ and $\theta_1$ using the constraint equations of the manipulator:

$$\begin{cases} A_{2x} + \rho_2 c_2 - \rho_1 c_1 - d_1 \cos(\alpha) = 0 \\ \rho_2 s_2 - \rho_1 s_1 - d_1 \sin(\alpha) = 0 \\ A_{3x} + \rho_3 c_3 - \rho_1 c_1 - d_3 \cos(\alpha + \beta) = 0 \\ A_{3y} + \rho_3 s_3 - \rho_1 s_1 - d_3 \sin(\alpha + \beta) = 0, \end{cases} \quad (8)$$

The first (resp. last) two equations make it possible to express $\rho_2$ (resp. $\rho_3$) as function of $\rho_1$, $\alpha$ and $\theta_1$. Then, $c_2$ and $s_2$ (resp. $c_3$ and $s_3$) are calculated as function of $\rho_1$, $\alpha$ and $\theta_1$ from the first (resp. last) two equations of (8) and their expressions are input in Eq. (7), which, now, depend only on $\rho_1$, $\alpha$ and $\theta_1$.

***Step 2:*** We fix a value for $\rho_1$, so Eq. (7) depends now only on $\alpha$ and $\theta_1$. By varying $\alpha$ or $\theta_1$, we compute the roots of the equation, to obtain the singular configurations ($\alpha_s$, $\theta_{1s}$) for a fixed $\rho_{1s}$.

***Step 3:*** For every singular configuration computed in the second step of the approach, we calculate the corresponding values $\rho_{2s}$ and $\rho_{3s}$ using equation system (9). We have thus the singular configurations curves in a slice of the joint space ($\rho_2$, $\rho_3$) for a fixed value of $\rho_1$:

$$\begin{cases} \rho_2 = \sqrt{\left(-A_{2x} + \rho_1 \cos(\theta_1) + d_1 \cos(\alpha)\right)^2 + \left(\rho_1 \sin(\theta_1) + d_1 \sin(\alpha)\right)^2} \\ \rho_3 = \sqrt{\left(A_{3x} - \rho_1 \cos(\theta_1) - d_3 \cos(\alpha + \beta)\right)^2 + \left(A_{3y} - \rho_1 \sin(\theta_1) - d_3 \sin(\alpha + \beta)\right)^2} \end{cases} \quad (9)$$

Figure 3 shows a slice of the joint space singular configurations for $\rho_1$=17 obtained for the same 3-R*P*R manipulator used by several authors [1,8,9], which has the following geometric parameters: $A_1$= (0, 0), $A_2$= (15.91,0), $A_3$ = (0, 10), $d_1$= 17.04, $d_2$= 16.54 and $d_3$ = 20.84 in an arbitrary length unit. We refer only to this manipulator in this paper in order to illustrate our work.

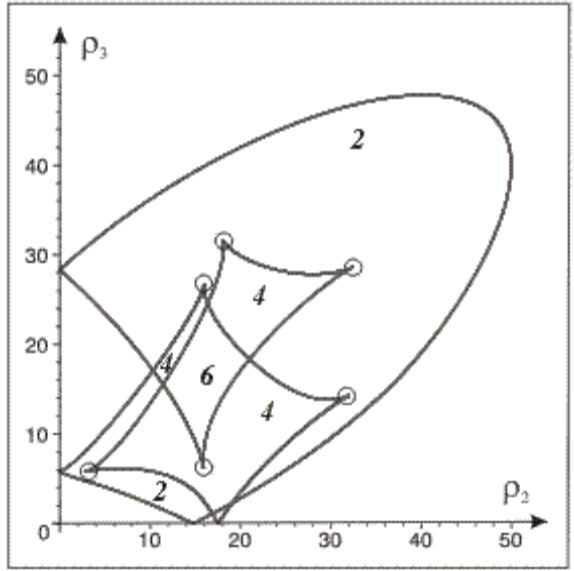

*Fig 3.   Singular curves in (α, θ₁) for ρ₁=17. Number of assembly modes is displayed in each region*

Figure 3 shows the singular curves in the joint space slice ($\rho_2$, $\rho_3$) defined by $\rho_1$=17. These curves give rise to several regions with 2, 4 or 6 direct kinematic solutions. The six points pinpointed with circles are cusp points, where three direct kinematic solutions coincide.

## 4. Determination of the Cusp Points

### 4.1 Existence condition of cusp points

For serial 3-DOF manipulators, the cusp points can be determined by deriving the condition under which the inverse kinematics polynomial admits three identical roots[13]. However this approach is much more complicated when applied to the direct kinematics polynomial of 3-R<u>P</u>R manipulators because this polynomial is of degree 6.

An interesting alternative approach was proposed in McAree's study[9] by writing the condition under which the manipulator loses first and second order constraints. The resulting condition for triple coalescence of assembly modes was shown to take the following form:

$$\mathbf{v}^T \left[ u_1 \frac{\partial^2 \Gamma_1}{\partial \boldsymbol{\theta}^2} + u_2 \frac{\partial^2 \Gamma_2}{\partial \boldsymbol{\theta}^2} + u_3 \frac{\partial^2 \Gamma_3}{\partial \boldsymbol{\theta}^2} \right] \mathbf{v} = 0, \qquad (10)$$

where **v** is a unit vector in the right kernel of matrix $\partial \boldsymbol{\Gamma}/\partial \boldsymbol{\theta}$, and $u_1$, $u_2$, $u_3$ are the three components of the unit vector **u** that spans the left kernel. Vectors **u** and **v** can be chosen in the set of nonzero rows and columns of the adjoint of matrix $\partial\Gamma/\partial\theta$ (i.e. the matrix of cofactors of the transpose of $\partial\Gamma/\partial\theta$), respectively.

Calculating the adjoint of $\partial\Gamma/\partial\theta$ from Eq. (5) yields:

$$\text{adj}\left(\frac{\partial\Gamma}{\partial\theta}\right) = \begin{pmatrix} k_1k_2 & -k_2k_5 & k_3k_5 \\ k_3k_4 & k_2k_6 & k_3k_6 \\ -k_1k_4 & k_4k_5 & k_1k_6 \end{pmatrix} \quad (11)$$

where

$$\begin{aligned}
k_1 &= 2\rho_2\left((A_{3x}-A_{2x})s_2 + \rho_3 s_{23} - A_{3y}c_2\right) & k_4 &= 2\rho_1\left(\rho_3 s_{13} + A_{3x}s_1 - A_{3y}c_1\right) \\
k_2 &= -2\rho_3\left(\rho_1 s_{13} + A_{3x}s_3 - A_{3y}c_3\right) & k_5 &= -2\rho_2\left(\rho_1 s_{12} + A_{2x}s_2\right) \\
k_3 &= -2\rho_3\left((A_{3x}-A_{2x})s_3 + \rho_2 s_{23} - A_{3y}c_3\right) & k_6 &= 2\rho_1\left(\rho_2 s_{12} + A_{2x}s_1\right)
\end{aligned} \quad (12)$$

Taking **u** (resp. **v**) as the first row (resp. column) of (11), Eq. (10) can be written as

$$\begin{pmatrix} k_1k_2 & k_3k_4 & -k_1k_4 \end{pmatrix} \left( k_1k_2 \frac{\partial^2\Gamma_1}{\partial\theta^2} - k_2k_5 \frac{\partial^2\Gamma_2}{\partial\theta^2} + k_3k_5 \frac{\partial^2\Gamma_3}{\partial\theta^2} \right) \begin{pmatrix} k_1k_2 \\ k_3k_4 \\ -k_1k_4 \end{pmatrix} = 0 \quad (13)$$

where

$$\frac{\partial^2\Gamma_1}{\partial\theta^2} = 2\begin{bmatrix} \rho_1(A_{2x}c_1 + \rho_2 c_{21}) & -\rho_1\rho_2 c_{21} & 0 \\ -\rho_1\rho_2 c_{21} & -\rho_2(A_{2x}c_2 - \rho_1 c_{21}) & 0 \\ 0 & 0 & 0 \end{bmatrix}$$

$$\frac{\partial^2\Gamma_2}{\partial\theta^2} = 2\begin{bmatrix} 0 & 0 & 0 \\ 0 & \begin{array}{c}-\rho_2((A_{2x}-A_{3x})c_2 \\ -\rho_3 c_{23} - A_{3y}s_2)\end{array} & -\rho_2\rho_3 c_{23} \\ 0 & -\rho_2\rho_3 c_{23} & \begin{array}{c}\rho_3((A_{2x}-A_{3x})c_3 \\ +\rho_2 c_{23} - A_{3y}s_3)\end{array} \end{bmatrix} \quad (14)$$

$$\frac{\partial^2\Gamma_3}{\partial\theta^2} = 2\begin{bmatrix} \rho_1(A_{3x}c_1 + \rho_3 c_{31} + A_{3y}s_1) & 0 & -\rho_1\rho_3 c_{31} \\ 0 & 0 & 0 \\ -\rho_1\rho_3 c_{31} & 0 & \rho_3(\rho_1 c_{31} - A_{3x}c_3 - A_{3y}s_3) \end{bmatrix}$$

In McAree's study[9], Eq. (13) was not expanded. McAree[9] noted that the expansion of this equation was too complicated to yield any real insight.

We have developed an algorithm to solve this equation for any 3-R*P*R manipulator and we have implemented it in Maple. This algorithm detects all the cusp points inside the joint space of any 3-R*P*R manipulators and computes their coordinates. We present it in next section.

**4.2 Algorithm for calculating cusp points**

The presence of cusp points allows the *3-RPR* manipulator to undertake non-singular assembly mode changing trajectories; these special trajectories can be executed by encircling a cusp point. In their study[9], the authors stated that cusp points are pernicious and should be avoided or designed out by judicious dimensioning.

The configuration of the *3-RPR* manipulator is given by six parameters: the three rod lengths ($\rho_1$, $\rho_2$, $\rho_3$), and the platform position variables ($\theta_1$, $\theta_2$, $\theta_3$). Only three of these parameters are independent. In order to reduce the dimension of the problem, McAree[9] shows that it is possible to consider two-dimensional slices of the configuration space by fixing one of the leg rod lengths.

By doing so, the manipulator configuration can be fully defined by only two parameters. For example, for a fixed value of $\rho_1$, a configuration may be fully defined by either ($\alpha, \theta_1$) or ($\rho_2, \rho_3$). Note that in the first case, the configuration is defined in the output space by the position and the orientation of the moving platform ($\rho_1$ and $\theta_1$ define the position of $B_1$ in the plane and $\alpha$ defines the orientation of the moving platform in the plane). In the second case, the configuration is defined in the joint space by the lengths of the three leg rods.

In our work, we have always taken $\rho_1$ as the fixed parameter. After fixing the value of $\rho_1$, we first calculate the singularity curves in ($\rho_2, \rho_3$), and then we compute all the cusp points of this two-dimensional slice.

*4.2.1 Algorithm*

If we consider equation (7), we notice that it is a function of ($\theta_1$, $\theta_2$, $\theta_3$). The existence condition of cusp points (13) is a function of ($\rho_1, \rho_2, \rho_3$) and ($\theta_1, \theta_2, \theta_3$). Our first goal is to establish an equation, which is a function of ($\rho_1, \alpha, \theta_1$), and then to solve it to obtain the coordinates of the cusp points. Thus, we first derive a set of equations from the geometry of the manipulator:

$$\begin{cases} \cos(\theta_2) = \dfrac{-A_{2x} + \rho_1 \cos(\theta_1) + d_1 \cos(\alpha)}{\rho_2} \\ \cos(\theta_3) = \dfrac{-A_{3x} + \rho_1 \cos(\theta_1) + d_3 \cos(\alpha + \beta)}{\rho_3} \\ \sin(\theta_2) = \dfrac{\rho_1 \sin(\theta_1) + d_1 \sin(\alpha)}{\rho_2} \\ \sin(\theta_3) = \dfrac{-A_{3y} + \rho_1 \sin(\theta_1) + d_3 \sin(\alpha + \beta)}{\rho_3} \end{cases} \qquad (15)$$

The algorithm for detecting cusp points is implemented in MAPLE; its steps are presented below:

1. First, the expression of $\cos(\theta_2)$, $\cos(\theta_3)$, $\sin(\theta_2)$ and $\sin(\theta_3)$ in (15) are substituted into the singularity equations (7). Then, $\sin(\alpha)$ and $\cos(\alpha)$ (resp. $\cos(\theta_1)$ and $\sin(\theta_1)$) are written as function of $\tan(\alpha/2)$ (resp. of $\tan(\theta_1/2)$). As a consequence we obtain an equation of the form:

$$F_1(\rho_1, t, t_1) = 0 \qquad (16)$$

where $t = \tan(\alpha)$ and $t_1 = \tan(\theta_1)$.

2. Then, the expression of $\cos(\theta_2)$, $\cos(\theta_3)$, $\sin(\theta_2)$ and $\sin(\theta_3)$ in (15) are substituted into equations (12) and (14). Applying then the tan-half substitution as above yields an equation of the form:

$$E_1(\rho_1, t, t_1) = 0 \qquad (17)$$

So, we notice that the two equations (7) and (13) are written now as function of three parameters only.

3. We fix now $\rho_1$, and we input the manipulator parameters $d_1$, $d_2$, $d_3$, $A_{2x}$, $A_{3x}$ and $A_{3y}$. We have noticed that the direct substitution of the real values of $\sin(\beta)$ and $\cos(\beta)$ into equations (16) and (17) make the equations resolution very complicated in the following steps. Thus, we write $\sin(\beta)$ and $\cos(\beta)$ as a function of an intermediate parameter $h$, which is the altitude of the moving triangle.

4. The Maple *resultant* function is used to eliminate $t = \tan(\alpha)$ from the two equations (16) and (17). The resulting equation is a polynomial of degree 96 in $t_1 = \tan(\theta_1)$, which can be factored as follows:

$$P_1^{a_1} P_2^{a_2} P_3^{a_3} ... P_{n-1}^{a_{n-1}} P_n^{a_n} Q = 0 \qquad (18)$$

where $Q$ is a $24^{\text{th}}$-order univariate polynomial in $t_1$ and $P_1, P_2,...,P_n$ are quadratic and quartic polynomials in $t_1$. Note that the factor form could not be obtained without the intermediate parameter $h$.

5. We input the parameter $h$ value. We solve equation (18). Each real root $t_{1i}$ is back-substituted into (16), which is then solved for $t$. For every $t_{1i}$, we obtain different values for $t_{ij}$. Finally, we get a number of solution couples ($t_{ij}, t_{1i}$).

6. We substitute the values of each solution couple ($t_{ij}, t_{1i}$) into (17), and we keep only those solutions that satisfy this equation.

The solutions ($t_{ij}, t_{1i}$) kept in the last step should give the coordinates of the cusp points. To verify this, we calculate the direct kinematic solutions for each solution ($t_{ij}, t_{1i}$). In many instances, we have found that some solutions do not yield three coincident solutions. This means that they are not associated with cusp points. So we reject them and we keep only those solutions that give three coincident direct kinematic solutions. These couples are the coordinates of the cusp points, we call them ($\alpha_{ij}, \theta_{1i}$)$_{\text{cusp}}$.

After running our algorithm hundreds of times, we have noticed that in each case all cusp points were determined by the $24^{\text{th}}$-order polynomial $Q$ of equation (18), that is, all remaining factors provided spurious solutions. Thus, we may conjecture that the cusp points are determined by $Q$, although we have no mathematical proof for this fact. All real roots of $Q$ are the cusp points.

With this conjecture, our algorithm simplifies significantly because instead of solving (18) (a 96$^{th}$ order polynomial), we just have to solve polynomial $Q$ (a 24$^{th}$ order polynomial).

To implement this result in the algorithm, we must change steps 5 and 6 into the following steps 5' and 6', and eliminate step 7:

5'. We input the real value of parameter $h$. We solve the polynomial $Q$. We substitute every real root $\theta_{1i}$ of $Q$ into equation (16), and solve it for $\tan(\alpha/2)$. For every $\theta_{1i}$, we obtain different values $\alpha_{ij}$. Finally, we get a number of couples $(\alpha_{ij}, \theta_{1i})$.

6'. We substitute the values of each couple $(\alpha_{ij}, \theta_{1i})$ into equation (17). The couples that satisfy this equation are the coordinates of the cusp points. We call them $(\alpha_{ij}, \theta_{1i})_{cusp}$.

Finally, to obtain the coordinates of the cusp points in the joint space $(\rho_1, \rho_2, \rho_3)$, we use the system of equations (9) computed from the geometry of the manipulator. We then obtain the cusp points in a slice of the joint space for a fixed value of $\rho_1$.

In step 7 of the algorithm, we have noticed that the existence condition of cusps generates solutions that do not provide triple direct kinematic solutions. This confirms the statement of McAree[9] that the cusp existence condition that he established is a sufficient condition but not a necessary one.

### 4.2.2 *Algorithm Execution*

The algorithm execution time slightly depends on the value of $\rho_1$ and of the *3-RPR* manipulator parameters. It highly depends on the number of digits required for the calculation. For 90 digits (which is necessary to guaranty a good accuracy), it is about two minutes on a computer equipped with a 3GHz-Pentium 4 with 512 Mo of Ram.

We present the results of an execution of the algorithm, for the same *3-RPR* parallel manipulator introduced in section 3. For the same fixed value $\rho_1=14.98$, as the one used in previous papers[8,9], the algorithm detects six cusp points instead of five identified in the paper[9]. Figure 4 shows the singular curves of the 3-*RPR* manipulator for $\rho_1=14.98$, and the six cusp points pinpointed with circles. The sixth point missed by McAree[9] is the point A, it is circled with bold lines and in-boxed in a separate view. The zoomed view shows that it is really a cusp point.

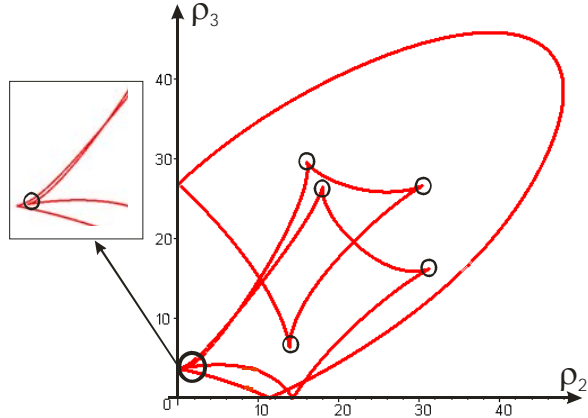

Fig 4. Singular curves and cusp points in a slice of the 3-RPR manipulator joint space $(\rho_2, \rho_3)$ for $\rho_1=14.98$.

The coordinates of the six cusp points are given in the table 1 below

|        | $\alpha$     | $\theta_1$   | $\rho_2$ | $\rho_3$ |
|--------|--------------|--------------|----------|----------|
| Cusp A | 50.67 deg    | -69.12 deg   | 0.84     | 3.77     |
| Cusp B | -2.59 deg    | 177.32 deg   | 13.85    | 6.26     |
| Cusp C | -122.89 deg  | 114.05 deg   | 31.27    | 16.17    |
| Cusp D | 57.48 deg    | 133.77 deg   | 30.44    | 26.61    |
| Cusp E | -0.59 deg    | 15.46 deg    | 16.02    | 29.56    |
| Cusp F | 170.37 deg   | -10.65 deg   | 17.98    | 26.44    |

Tab 1. Coordinates of the six cusp points for $\rho_1=14.98$.

5. **Descriptive Analysis**

In this section, the singular curves are analysed in several slices of the joint space for the manipulator studied in papers[8,9] and whose geometric parameters were given in section 3. Figure 5 depicts the singular curves for increasing values of $\rho_1$ and shows that the number of cusp points is not the same for all slices as we may have 0, 2, 4, 6 or 8 cusp points. In Fig. 5, regions with two assembly modes (resp. four, six) are filled in light grey (resp. in dark grey, in black). Zero-cusp slices are obtained for very small values of $\rho_1$ only ($\rho_1=0.05$ in Fig. 5), where the singular curves are made of two separate closed curves that define only one small region with two assembly modes and a large void (note, the two curves are so close that the region cannot be seen in Fig. 5). When $\rho_1$ is increased, two cusp points appear, the void gets smaller and a four-solution region appears ($\rho_1$=2). Then two more cusp points appear, defining one more four-solution region ($\rho_1$=2.8). We have six cusp points and a small void at $\rho_1$=6; for $\rho_1$=8, 10, 12, 14, 16, 18, 20, 24 and 26, we have always six cusp points but the void is replaced with a six-solution region. Eight cusp points exist in a small vicinity of $\rho_1$=27. Then two cusp points and the six-solution region disappear ($\rho_1$=29). Finally, the number of cusp points stabilizes to four, defining one central four-solution region surrounded by a two-solution region ($\rho_1$=31, …). Interestingly, this last pattern is very similar to the one often observed in a cross-section of the workspace of 3-R serial manipulators[19]. However, serial manipulators feature the same pattern in all cross-sections (the sections which passes through the first revolute joint axis), and variation in the number of cusp points arises only from a modification of the manipulator geometry.

The above analysis shows that the joint space topology of 3-R<u>P</u>R manipulators is very complicated. Contrary to serial manipulators, the shape of the singular curves and the number of cusps points depend on which slice is chosen in the joint space. Thus, planning trajectories is not easy. However, we have noticed that the pattern stabilizes for sufficiently large values of $\rho_1$.

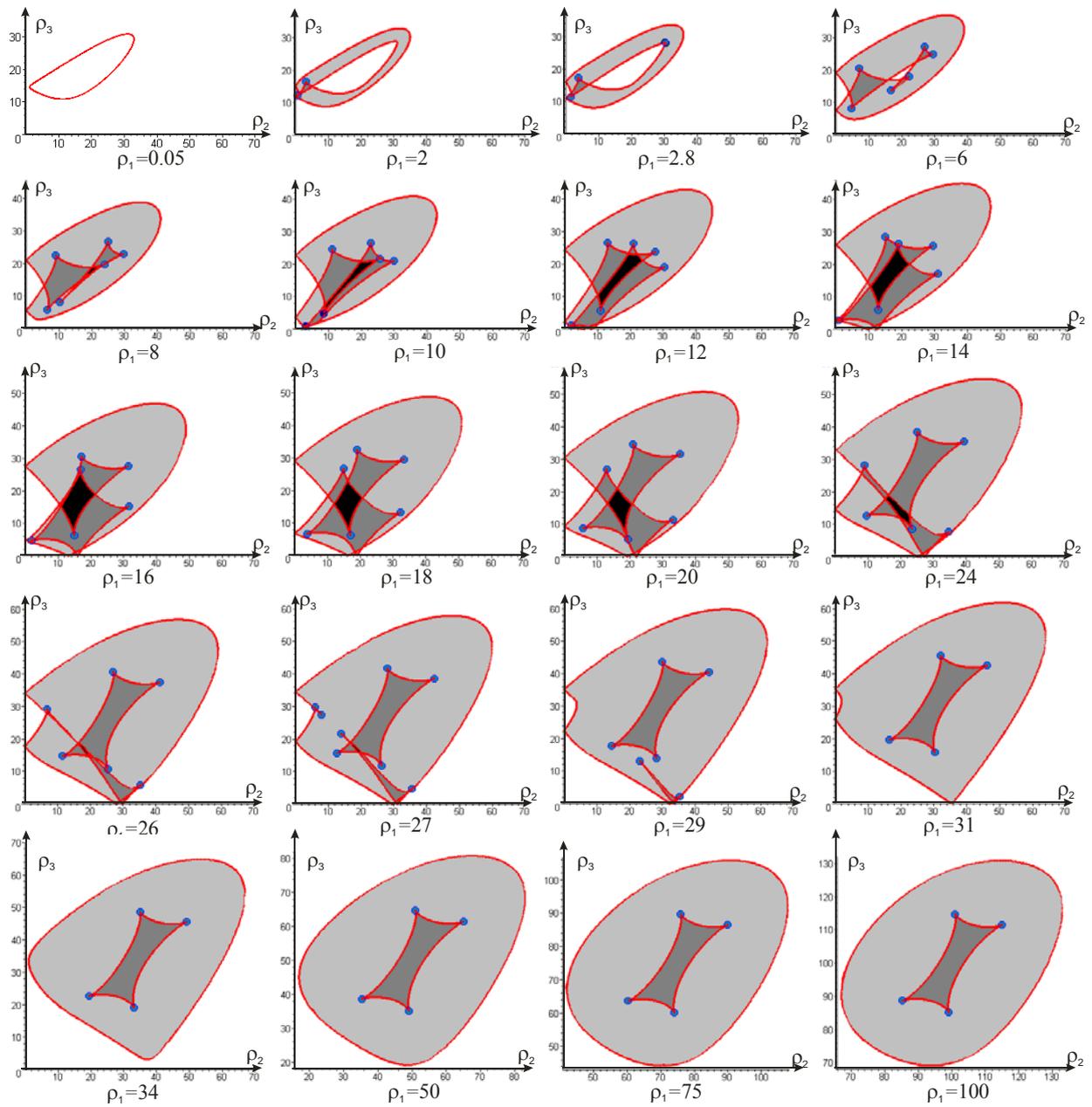

*Fig 5. Singular curve patterns for increasing values of $\rho_1$.*

Figure 6 represents the singularities in the joint space of the manipulator studied when $\rho_1$ varies from 0 to 50. To obtain this surface, we have imported the solutions obtained in step 4 into a CAD software, and we have meshed them together.

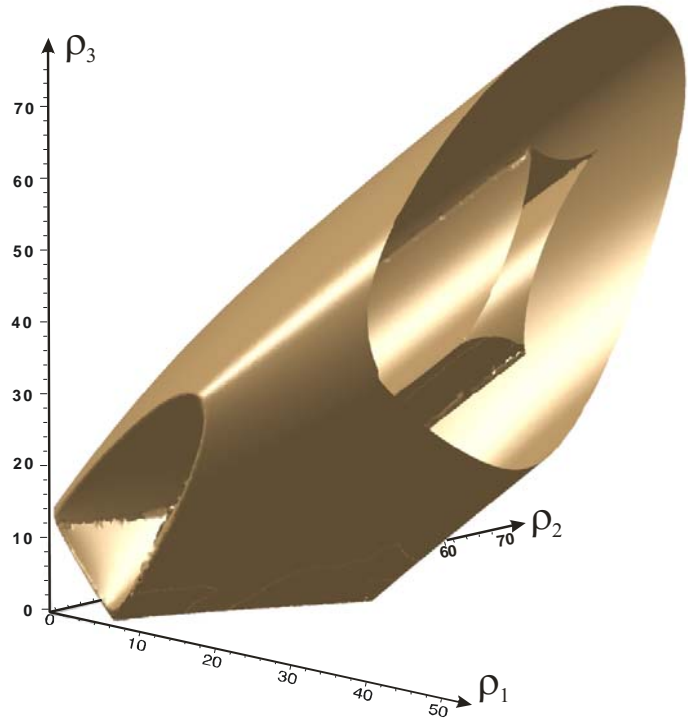

*Fig 6. Joint space singularity surfaces of the 3-RPR manipulator studied when $\rho_1$ varies from 0 to 50.*

The surface depicted in Figure 6 is of interest:

***i.*** for planning trajectories in the joint space because it shows clearly the joint space regions that are free of singularities.

***ii.*** for manipulator design, because it offers a tool for defining the values of the joint limits such that the joint space is a singularity-free box.

We can see clearly in figure 6 how the number of cusp points and the topology of the slices stabilize for $\rho_1>31$.

This feature has been observed in many other manipulator geometries. For example, the 3-RPR manipulator defined by the following geometric parameters: $A_1$= (0,0), $A_2$= (3,0), $A_3$ = (1.1,2.7), $d_1$= 1.3, $d_2$= 0.9 and $d_3$ = 0.4 in an arbitrary length unit, has a constant pattern as soon as $\rho_1>5$ (Fig. 7). Note that, in contrast with the stabilized pattern obtained for the preceding manipulator, this pattern features a large void.

Most of the research work on parallel manipulators has focused on the analysis and optimization of the workspace. If the workspace is useful for manipulator design, the analysis of singular curve patterns in the joint space can be used as a complementary tool to compare several manipulator geometries.

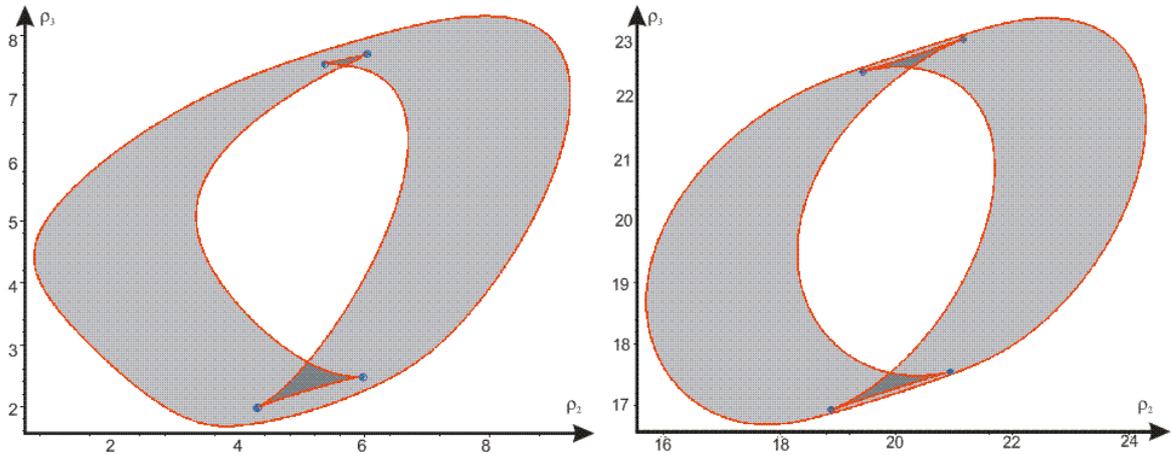

Fig 7.  Stabilized singular curve patterns for another manipulator geometry plotted for $\rho_1=5$ (left) and $\rho_1=20$ (right).

## 6. Conclusions

A procedure for computing joint space singularities of *3-RPR* parallel manipulators has been presented in this paper. Then, the existence condition of cusp points defined by McAree[9] has been reviewed and exploited. An algorithm able to detect and to compute cusp points inside any section of the joint space of any *3-RPR* parallel manipulator has been established. To the best of the authors' knowledge, such an algorithm had never been proposed before. The algorithm results in a 96th degree univariate polynomial that can be put in a factored form. We have showed with intensive numerical experiments that the coordinates of the cusp points are the real roots of a 24th degree univariate polynomial, which is one of the factors of the 96th polynomial. Finally, a descriptive analysis of the singular curves in slices of the joint space of 3-*RPR* parallel manipulators has been conducted, with a special focus on the cusp points on these singular curves. It has been shown that, contrary to what arises in serial manipulators where any cross section of the workspace exhibits the same pattern of singular curves and cusp points, this pattern depends on the choice of the slice in the joint space for a given 3-*RPR* parallel manipulator. On the other hand, we have noticed that it is possible to have a constant pattern by adjusting the joint limits. It has been observed that the maximum number of cusps depends on the manipulator geometry . No manipulators were found with more than eight cusp points. Most of the research work on parallel manipulators has been focused on the analysis and optimization of the workspace. This paper provides a complementary tool that helps better understand the topology of the joint space of parallel manipulators and finds applications in both design and trajectory planning.